\documentclass{article}
\usepackage{ijcai25}

\usepackage{times}
\usepackage{soul}
\usepackage{url}
\usepackage[hidelinks]{hyperref}
\usepackage[utf8]{inputenc}
\usepackage[small]{caption}
\usepackage{graphicx}
\usepackage{amsmath}
\usepackage{amsthm}
\usepackage{booktabs}
\usepackage[switch]{lineno}

\usepackage{times}
\usepackage{latexsym}

\usepackage{amsmath}
\usepackage{mathtools}

\usepackage{enumitem}
\usepackage{booktabs}
\usepackage{lipsum}
\usepackage{graphicx}
\usepackage{url}
\usepackage{tabularx}
\usepackage{longtable}
\usepackage[dvipsnames]{xcolor}
\usepackage{dsfont}

\usepackage[ruled,linesnumbered,noend]{algorithm2e}

\usepackage[mathscr]{euscript}

\newcommand\tldrDone[1]{}

\usepackage{graphicx}
\usepackage{subcaption}

\newcommand{\detect}{\mathtt{detect}}

\newcommand{\len}[1]{\mathtt{len}(#1)}

\newcommand{\R}{\mathbb{R}}

\newcommand{\E}[2][]{\mathbb{E}_{\ifx &#1& \else #1 \fi}\left[#2\right]}

\renewcommand{\P}[1]{\mathbb{P}\(#1\)}

\newcommand{\Ep}{\mathbb{E}}

\renewcommand{\P}{\mathbb{P}}

\newcommand{\est}[1]{\widehat{#1}}

\newcommand{\mc}[1]{\mathcal{#1}}

\newcommand{\indic}[1]{\mathbf{1}\!\left\{#1\right\}} 

\newcommand{\defeq}{:=}

\definecolor{innerboxcolor}{rgb}{.9,.95,1}
\definecolor{outerlinecolor}{rgb}{.6,0,.2}
\definecolor{shcolor}{RGB}{27, 87, 14}
\definecolor{rckcolor}{RGB}{0,0,255}

\newcommand{\wt}[1]{\widetilde{#1}}

\usepackage{multirow}
\usepackage{amsmath}
\usepackage{todonotes}
\usepackage[
	lambda,
	operators,
	advantage,
	sets,
	adversary,
	landau,
	probability,
	notions,	
	logic,
	ff,
	mm,
	primitives,
	events,
	complexity,
	asymptotics,
	keys]{cryptocode}

\newtheorem{definition}{Definition}

\title{Entropy-Guided Watermarking for LLMs: \\A Test-Time Framework for Robust and Traceable Text Generation}

\author{
Shizhan Cai$^1$\and
Liang Ding$^{2}$\and
Dacheng Tao$^1$\\
\affiliations
$^1$Nanyang Technological University\quad $^2$University of Sydney\\
\emails
shizhan.cai@ntu.edu.sg,
liangding.liam@gmail.com,
dacheng.tao@gmail.com
}

\begin{document}
\maketitle
\begin{abstract}
The rapid development of Large Language Models (LLMs) has intensified concerns about content traceability and potential misuse. Existing watermarking schemes for sampled text often face trade-offs between maintaining text quality and ensuring robust detection against various attacks. To address these issues, we propose a novel watermarking scheme that improves both detectability and text quality by introducing a cumulative watermark entropy threshold. Our approach is compatible with and generalizes existing sampling functions, enhancing adaptability. Experimental results across multiple LLMs show that our scheme significantly outperforms existing methods, achieving over 80\% improvements on widely-used datasets, e.g., MATH and GSM8K, while maintaining high detection accuracy. The code will be released.

\end{abstract}

\section{Introduction}
Large Language Models (LLMs) have profoundly impacted our lives~\cite{javaheripi2023phi,team2024gemma,dubey2024llama,zhong2023can}. However, their widespread application has also introduced challenges, including the spread of misinformation and disputes over copyright. In this context, watermarking in LLMs has emerged as a critical countermeasure to enhance model accountability and combat misuse~\cite{kirchenbauer2023watermark,kuditipudi2023robust,guo-etal-2024-context-aware}. By embedding watermarks, LLM owners can better monitor model usage, safeguard intellectual property, and mitigate the risks of unauthorized distillation training on model outputs by making the process traceable.

\begin{figure}[t]
    \centering
    \includegraphics[width=0.94\columnwidth]{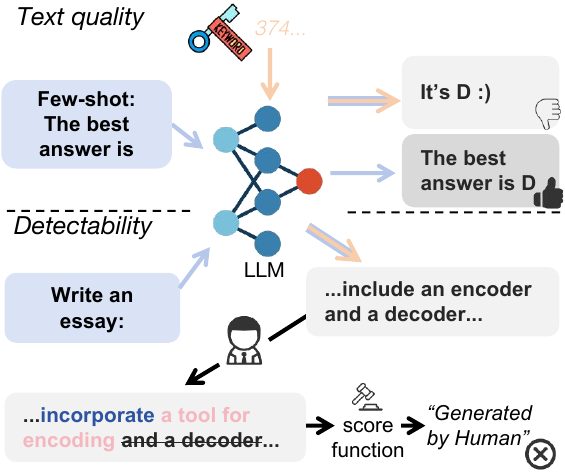}
    \caption{\textbf{Overview of the watermarking workflow during LLM sampling.} \textcolor{Apricot}{Apricot arrows} represent the implantation of the \textcolor{Apricot}{secret key}, while \textcolor{RoyalBlue}{blue arrows} illustrate the original LLM generation flow. The watermarking approach must balance two critical requirements: (1) \textbf{text quality}—ensuring that output text retains the same quality as non-watermarked text to preserve user experience; and (2) \textbf{detectability}—making the watermark reliably detectable, even when users modify the output. Existing schemes exhibit text quality degradation and weaknesses in detectability under adversarial attacks.}
    \label{introduction}
\end{figure}

An LLM watermarking scheme comprises two components: embedding and detection. Embedding can be performed during the logits or sampling phases. In the logits watermark~\cite{kirchenbauer2023watermark,hu2023unbiased}, a hash function using previous tokens as seed is employed to partition the token vocabulary into a whitelist and blacklist, with biases applied to whitelist logits. Detection then relies on statistical scoring. In contrast to logit-based watermarks, sampling-based watermarking~\cite{kuditipudi2023robust,christ2023undetectable} aligns watermarking with the original distribution by embedding random keys during sampling, enabling detection through these keys. A key advantage of sampling watermarking is its ability to preserve the original text distribution.

We argue that existing sampling watermarks face a trade-off between maintaining text quality and ensuring robust detection under various attacks. For example, the scheme proposed by~\cite{kuditipudi2023robust} provides robust detection by measuring the distance between the watermark text and the given secret key. However, its reliance on a fixed key space can degrade output quality, leading to irregular output for few-shot prompts or templates (as illustrated in Figure~\ref{introduction}), which shows how deviations from expected text patterns, e.g., inconsistencies in few-shot tasks (``The best answer is ...'' or deterministic outputs (e.g., argmax decoding), may reveal the presence of the watermark. Similarly, the scheme by~\cite{christ2023undetectable} which uses a score function for detection, struggles against text modification attacks (detailed in Section~\ref{detectability}). These limitations hinder both user experience and the effectiveness of detection. Thus, a refined scheme is needed that balances alignment with the original text distribution and robustness in detection.

In this paper, we propose a novel scheme designed to harness the detectability and text quality by introducing a threshold for the cumulative watermark entropy. Outputs remain unwatermarked below this threshold while exceeding text is watermarked using preceding tokens as a seed to generate a key. Our scheme adapts to scenarios like few-shot prompts by controlling the threshold to align with user templates, and ensures the consistency in argmax outputs by fixed seeds and keys. Meanwhile, we adapt the binary sampling used by~\cite{christ2023undetectable} to our scheme by constructing a new mapping. We theoretically and empirically prove the effectiveness of this mapping on the detection metric.

We systematically evaluate our scheme over various LLMs, demonstrating high detectability even under strong paraphrase attacks. Our scheme shows only a 10\% AUC drop, compared to a 60\% drop in ~\cite{christ2023undetectable}, highlighting significant robustness. While ensuring the detectability, our experiments demonstrate that, compared to previous scheme~\cite{kuditipudi2023robust}, our framework achieves an 80\% improvement on long answer QA datasets such as MATH~\cite{hendrycksmath2021}, GSM8K~\cite{cobbe2021training}. By grounding our approach in both rigorous theoretical analysis and empirical experimentation, our scheme achieves a robust balance between indistinguishability and detectability.
In short, our main \textbf{contributions} are as follows:
\begin{itemize}
    \item We propose a novel watermarking scheme by introducing an entropy threshold, and theoretically prove that it is indistinguishable. 
    \item Our method is compatible with both existing sampling functions and has demonstrated their effectiveness in terms of text quality and detectability, highlighting its generalization capability.
    \item Extensive experiments show that our scheme consistently outperforms the previous schemes by a large margin on long answer tasks over several LLMs while maintaining high detectability.
\end{itemize}

\section{Task Definitions}

\begin{figure*}[ht]
    \centering
    \includegraphics[width=\textwidth]{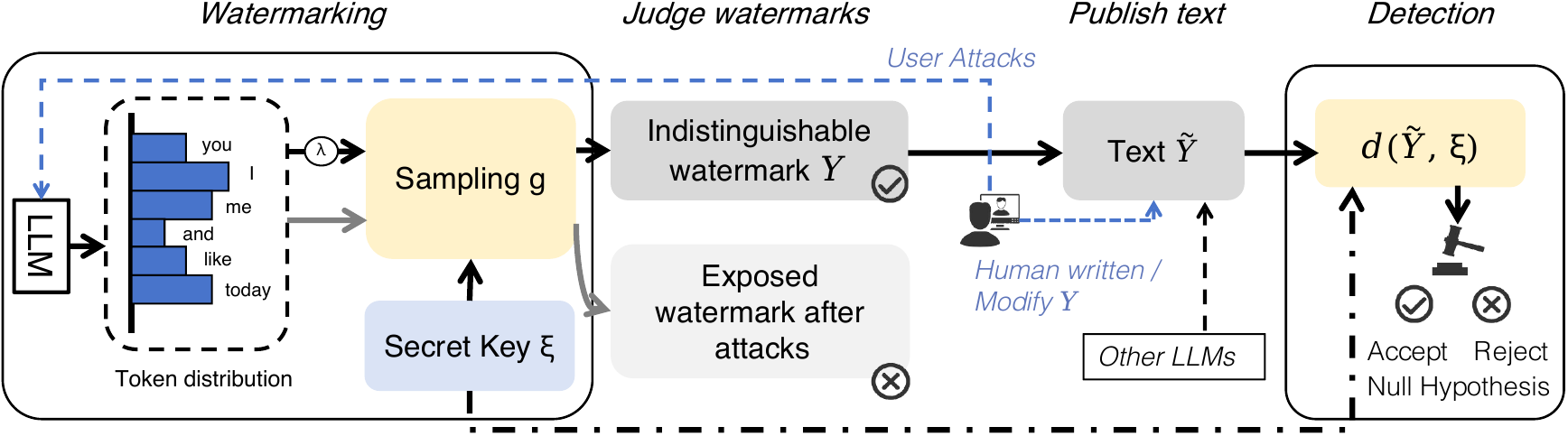}
    \caption{\textbf{Workflow of the proposed watermarking and detection algorithm.} The diagram illustrates the core steps of the watermarking process, including token distribution manipulation, secret key sampling, watermark embedding, and subsequent detection. Starting with a token distribution from the LLM, a secret key ($\xi$) is sampled to produce an indistinguishable watermark ($Y$), which is embedded into the generated text ($\tilde{Y}$). The watermark remains robust against various modification attacks, such as deleting, inserting, and substituting. Detection involves testing whether the text contains the watermark, either accepting or rejecting the null hypothesis.}
    \label{flowchart}
\end{figure*}

Let $\mc{V}$ represent the vocabulary set, and let $p \in \mc{V}^* \to \Delta(\mc{V})$ denoted the probability distribution outputted by LLM. The model maps user prompt $x\in \mc{V}^*$ and current prefix tokens $y_{:i-1}\in \mc{V}^*$ to a probability distribution over the vocabulary, where $p(y_i|x, y_{i-1})$ specifies the conditional distribution of the next token. Let $\Xi^*$ represent the space of the watermark key sequence. As the example of the student essay mentioned in the introduction, the analogy of watermark setting is as follows: 1. The student sends a prompt $x \in \mc{V}^*$ to the LLM. The LLM generates watermark essay $Y \in \mc{V}^*$, denoted by $Y = M_w(x, \xi)$, where $M_w$ denotes LLM with the watermark secret key sequence $\xi \in \Xi^*$. 2. The LLM owner shares the secret key $\xi$ with the teacher. 3. The student submits essay $\wt{Y} \in \mc{V}^*$, which may be either:(a) a watermark text $Y$ might have some modification; (b) an independent text of $Y$, for instances, the student writes his essay or the student use another LLM. 4. The teacher set null hypothesis: $\wt{Y}$ is independent of $\xi$ and detect function $\detect: \mc{V}^* \times \Xi^* \to \{-1,+1\}$, where negative label stands non-watermarked and positive label stands watermarked. By computing $\est{p}=\detect(\wt{Y}, \xi)$, the teacher chooses to reject the null hypothesis or not.

\subsection{High Entropy}
The entropy of a text measures the degree of uncertainty within its content. The text must exhibit an entropy higher than a certain threshold for a watermark to be effectively embedded without being easily detectable or reversible. If the entropy of the output text is low, which means the text is highly deterministic, any alterations to add a watermark make it easy to notice and ruin the original text's meaning. For example, given the prompt: \textit{The most famous Shakespeare's saying}, the best response is \textit{To be, or not to be}; in this case, it's meaningless to watermark the answer. Thus, we define watermark entropy as a text entropy criterion.

\begin{definition}
    Define watermark entropy $\alpha : \mc{V} \to \R$ by 
    \begin{align*}
        \alpha \defeq f(y)
    \end{align*}
\end{definition}

We defer the corresponding watermark entropies after introducing the sampling functions $g$ in the definition~\ref{sample function}

\section{Methodology}
\subsection{Watermarking Algorithm}
\label{watermark}
Indistinguishability is a key property for watermarks. Ideally, when a user makes many adaptive queries, it is infeasible to distinguish between the original and the watermarked models. However, the text may be corrupted since the watermark during sampling is highly dependent on the secret key $\xi$.

\begin{algorithm}
\caption{Watermarking algorithm}
\label{watermark_algorithm}
    \KwData{A prompt $x$, the secret kefy space $\Xi$ and the parameter $\lambda$}
    \KwResult{Watermarked text $Y$}
    $\text{Entropy} \gets 0$\\
    \For{$i \in 1,\ldots,m$}{
        $p_i \gets M(x, y_1, \ldots, y_{i-1})$\\
        \uIf{$\text{Entropy} < \lambda$}{
            Sample $y_i$ from $p_i$\\
            $\text{Entropy} \gets \text{Entropy} + \alpha(y_i)$\\
            \If{$\text{Entropy} \ge \lambda$}{
                Set $\{y_1, \dots, y_i\}$ as seed $r$\\
            }
        }
        \Else{
            $\xi(r)\sim \Xi$\\
            $y_i \gets g(\xi(r), p_i)$
        }
        }
\end{algorithm}

The scheme proposed in \cite{kuditipudi2023robust} introduces a secret key space $\Xi^n$ of fixed length $n$ to enhance watermarking. To manage multiple responses, they define a starting point $\tau$ over a uniform distribution $U[n]$ to vary the secret key sequence. This reduces the likelihood of key collision, with the expected number of generations $m$ for a collision being $O(\sqrt{n/m})$, analogous to the birthday problem (details in Appendix~\ref{birthday problem}). However, in practical use cases, such as brainstorming with repeated prompts, or deterministic queries like few-shot learning, the watermarked text may exhibit deviations. We refer to these deviations as user attacks. To address this, we adopt the strict indistinguishability definition from \cite{christ2023undetectable}, ensuring that even under polynomially many adaptive queries, responses remain indistinguishable from those of the original model. Figure~\ref{flowchart} demonstrates the process and key comparisons, highlighting the robustness of our method.

\begin{definition}
\label{indistinguishable}
    A watermarking model $M_w$ is \emph{indistinguishable} if calling polynomial-time distinguishers $D$ (A well algorithm that could distinguish text) for any parameter $\lambda$, we have
    $$
    \left|\P[D(M)=1]-\P[D(M_w{(\xi(\lambda))})=1]\right| \le \negl.
    $$
\end{definition}

To handle these attacks properly, we propose the watermarking algorithm~\ref{watermark_algorithm}. We sample text normally, i.e. without watermark implanted, until the text reaches out the $\lambda$ bits of watermark entropy. Then we set the whole block of previous tokens as the seed to generate the secret key $\xi$. 

We can prove that if we use $\alpha_i=1-p(y_i)$ as our watermark entropy, $M_w$ following the definition~\ref{indistinguishable} is indistinguishable. Suppose we have polynomial time queries $t=\text{poly}(\lambda)$. Let $r^{(1)}, r^{(2)}, \ldots r^{(t-1)}$ be the seed of responses $Y^{(1)}, Y^{(2)}, \ldots Y^{(t-1)}$. If the previous blocks are identical, the seeds are also equal so that the responses are same. In other case, Considering for some $k\in[t]$ the watermarking algorithm stops before collecting enough entropy, we let $r^{(k)}\coloneq \operatorname{None}$. Define set $B\coloneq\{r^{(1)}, r^{(2)}, \ldots r^{(t-1)}\}\setminus \{\operatorname{None}\}$.
For any $r^{(k)}=\operatorname{None}$, it's trivial to show the indistinguishability since the text is sampled from the original distribution. Then we show $\P[r^{(t)}\in B]\leq \text{negl}(\lambda)$, which means we have a negligible probability of colliding the key sequence. (The detailed proof is shown in appendix~\ref{indis_its}). 

Then we sample the token by the key $\xi$ and the token distribution $p_i$. The sampling algorithm $g$ is a way to combine them. Here is the definition of our function $g$:
\begin{definition}
\label{sample function}
Define sampling function $g:\Delta(\mc{V})\times\Xi\to\mc{V}$ by
$$
g\coloneq f(p(y), \xi(r)).
$$
\end{definition}
Specifically, there are two current sampling: inverse transform sampling (ITS)
\begin{equation}
\label{g_its}
g = \pi^{-1}(\min \{\pi(k): p(\{j: \pi(j) \leq \pi(k\}) \geq u\}),
\end{equation}
where $\pi$ is a random permutation and $u\in \xi(r)$, in \cite{kuditipudi2023robust}, or binary sampling (BS) in~\cite{christ2023undetectable}:
\begin{equation}
\label{g_bs}
g = E^{-1}\mathds{1}(E(p_i)\geq u), 
\end{equation}
where $E$ is the Huffman encoding as we defined in the appendix~\ref{huffman encoding} and $u\in \xi(r)$.

\subsection{Detection Algorithm}
\label{detection}
The other key property is detectability, which could ensure that our embedding watermark can be detected. In our task, the watermark detection is a binary classification. There are two important error rates: false positive rate (FPR) occurs when the detection mechanism fails to identify a watermark in content that is genuinely watermarked, and false negative rate (FNR) occurs when any text independent of the secret key is detected as watermarked. Denote total error $e=\operatorname{FNR}+\operatorname{FPR}$. Let $Y_i'$ be the nonwatermarked text and $Y \stackrel{d}{=} Y'$ be the watermarked text of length $m$. Let $\xi \in \Xi^*$ be a random variable that is independent of $Y'$. Define the set $\mc{V}_c$ by 
$\mc{V}_c \defeq \{y : p(y_i \mid y_{:i-1}) \geq \exp(-c)\}.$ Then \cite{kuditipudi2023robust} prove the total error rates of ITS: $e\geq \Ep\left[\exp(-cm\alpha(Y))\mathds{1}\{Y\in\mc{V}_c\}\right],$
where $\alpha = 1-p(y_i)$. This inequality implies the lower bound of the sum of the Type I and II errors will be large if the output text is likely deterministic. For example, when $c=0.1$, $p(y_i)\geq 0.95$, then $e\geq r$, where $r\to 1$. Then, it is impossible to test between any watermarked and non-watermarked. For our design watermark algorithm, we can easily control the watermark entropy by setting the parameter $\lambda$. For the lower entropy text, like the seed is still empty, we set the text as the negative label, which can reduce the error rate.

\begin{algorithm}
\caption{Detection algorithm}
\label{detect_algorithm}
    \KwData{string $y \in \mc{V}^*$; watermark key sequence $\xi \in \Xi^n$; cost $d$; resample size $T$}
    \KwResult{Detect p-value $\est{p} \in [0,1]$}
    \For{$t \in 0, 1,\dots,T$}{
        \uIf{$t=0$}{$\xi^{(0)}=\xi$\;}
        \Else{$\xi^{(t)} \sim \Delta(\Xi^n)$\;}
        \For{$i \in 1,\dots,\len{y}-k + 1$}{
        \For{$j \in 1,\dots,n$}{
            $y^i \leftarrow \{y_{i+\ell}\}_{\ell = 0}^{k-1}$,
            $\xi^j \leftarrow \{\xi^{(t)}_{(j+\ell) \% n}\}_{\ell=0}^{k-1}$\;
            $\phi^{t} \leftarrow \min\{ \phi^{t}, d(y^i,\xi^j)\}$\;
        }
    }
    }
    $\est{p} \leftarrow \frac{1}{T+1} \left(1 + \sum_{t=1}^T \indic{\phi_t \leq \phi(y,\xi)}\right)$\;
    \Return $\est{p}$\;
\end{algorithm}

To robustly detect the watermark, We follow the fine-grained detection algorithm~\ref{detect_algorithm} designed in \cite{kuditipudi2023robust}. To judge the watermarked text and non-watermarked text, the detector sets the null hypothesis that $\wt{Y}$ is not watermarked, i.e., that $\wt{Y}$ is independent of $\xi$. The detector uses the $\detect$ method to compute a $p$-value with respect to a test statistic $\phi: \mc{V}^* \times \Xi^* \to \R$ with a size $T$ resampling. For the statistic $\phi$, it is used to measure the distance between $\wt{Y}$ and any key $\xi\sim\Xi^*$. If $\wt{Y}$ is watermarked, $\phi$ will return a small value, e.g. 10e-3, since the $\wt{Y}$ is generated by the $\xi$. In the opposite, $\wt{Y}$ is independent with the original key $\xi$ or resampled key $\xi^{(t)}$. The statistic $\phi$ returns a random but large number.

Then to make the test statistic $\phi$ such that $\est{p}$ will typically be small if $\wt{Y}$ is watermarked. In particular, it needs a fine-grained metric over the key sequence and the response against the attacks. Here it comes out the definition of alignment cost:
\begin{definition}
Define cost $d:(\mc{V} \times \Xi)^* \to \R$:
$$
d\coloneq f(y,u)
$$
\end{definition}
which measures the quality of a match between a subsequence of the input text and a subsequence of the watermark key, and uses this to define $\phi$ as the minimum cost alignment between length $k$ subsequences of the text and key. For the inverse transform sampling, one way is to use the negative covariance $d(y,(u,\pi)) = -\sum_{i=1}^{\len{y}} (u_i - 1/2) \cdot (\eta(\pi_i(y_i)) - 1/2)$. \cite{kuditipudi2023robust} prove the effectiveness of this distance for the statistic $\phi$.

For the binary sampling, we show that the expectation of cost
\begin{equation}
\label{d_bs}
d(y,u) = -\sum_{i=1}^m (h(u_i) - 1/2) \cdot (\eta(y_i) - 1/2)
\end{equation}
has a gap between the resample keys $\xi^{(t)}$ and the secret key $\xi$, where $m={\len{y}}$, if the text $Y = M_w(\xi, x)$ is watermarked.
$$
\mathbb{E}\left[d(Y, \xi^{(t)})-d(Y, \xi)\mid Y\right]=m\operatorname{Var}(\eta(Y))\alpha(Y)
$$
where $h$ maps the secret keys for $y_i$ to a random number in $[0, 1]$. Here is the detailed construction of $h:\Xi^*\to\mathbb{R}$. For token $y_i$, let $l={\operatorname{len}(E(y_i))}$, where $E$ is Huffman encoding. Denote the secret key sequence for this token $\{u_j\}_{j=0}^{l-1}$, we have
$$
h(\{u_j\}_{j=0}^{l-1})\coloneq\eta(E^{-1}(\{\mathds{1}(u_j>\frac{1}{2})\}_{j=0}^{l-1}))/N,
$$
where $N$ is the length of the vocabulary set and $\eta(i)=\frac{i}{N}$. (The detailed proof is shown in the appendix~\ref{binary construction}). Then to ensure that the resampled key has a low probability of having a lower distance than the secret key. Let $Y_{i:i+k-1}$ be a substring of $Y$ of length $k$. For any block of size $k$, we show that
\begin{align*}
    & \P\left(d(Y_{i+1:i+k},\xi_{j+1:j+k}^{(t)}) \leq d(Y_{i:i+k-1},\xi_{i+1:i+k})\right)\\ 
    & \leq 2\exp\left(-m \operatorname{Var}(\eta(Y))^2 \alpha^2/2\right).
\end{align*}
The detailed deviation is shown in the appendix~\ref{proof_p}

\section{Evaluation}
We test two metrics for evaluating watermarking schemes: (a) quality and (b) detectability.

\subsection{Models}
We evaluate on 4 light LLMs: (1) Llama-3.2-1B~\cite{dubey2024llama} (2) OPT-1.3B~\cite{zhang2022opt} (3) Gemma-2B~\cite{team2024gemma} (4) phi-2B~\cite{javaheripi2023phi}.

\subsection{Detectability}
\label{detectability}
\begin{figure*}[t]
    \centering
    \includegraphics[width=\textwidth]{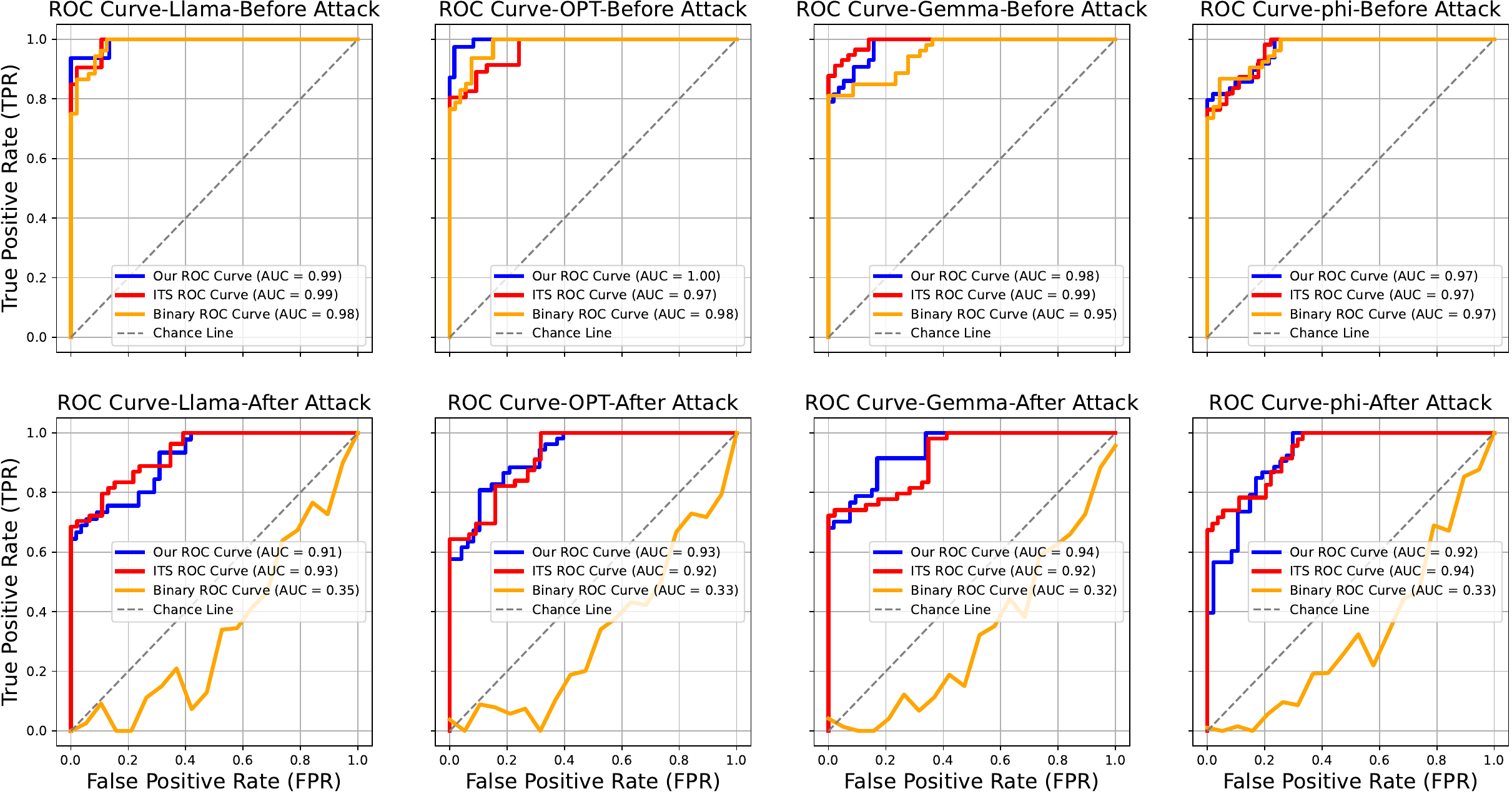}
    \caption{\textbf{ROC curves for our, ITS, and Binary scheme under different attack conditions.} The first row shows ROC curves for watermarked text before attacks, while the second row illustrates the impact of paraphrasing attacks on the same text. Each subplot corresponds to a specific model (Llama, OPT, Gemma, phi). Our scheme demonstrates superior performance, achieving high AUC values both before and after attacks, with minimal degradation in classification ability compared to ITS and Binary schemes. In contrast, the Binary scheme shows significant vulnerability, with AUC values dropping below 0.35 post-attack, highlighting its limited robustness in adversarial scenarios.}
    \label{roc_curve}
\end{figure*}

We empirically compare our scheme with the original implementation in \cite{kuditipudi2023robust} and \cite{christ2023undetectable} via the models in the list. We use the abbreviations "ITS" and "BS," respectively. For ITS scheme, we use $d(y, u)=-\operatorname{Cov}(\eta(\pi(y)), u)$. In the original BS scheme, they used a score function as the statistic. To show detection effectiveness, we use binary sampling in our scheme. For the cost $d$ of binary sampling, we use the adapted $d(y,u) = -\sum_{i=1}^m (h(u_i) - 1/2) \cdot (\eta(y_i) - 1/2)$. We generate 100 watermarked text continuations of prompts sampled from the C4 dataset~\cite{raffel2020exploring} from the four models mentioned above. 

We tested three attack methods—Basic Attack, Translation Attack, and Paraphrase Attack—on the Binary Scheme (BS) by evaluating the total error $e$. The results of these attacks are summarized in Table~\ref{attack comparison}. As shown in Figure~\ref{roc_curve}, before any attack (first row), all schemes, including our proposed method, ITS, and Binary Scheme, achieved strong performance with AUC values exceeding 0.95. However, under adversarial conditions, particularly the paraphrase attack, significant differences emerged.

The paraphrase attack, implemented as the most effective strategy, showed the most profound impact. For example, focusing on the Llama model, our scheme demonstrated robustness with an AUC of 0.91 post-attack, compared to the Binary Scheme, whose AUC drastically dropped from 0.98 to 0.35, highlighting its vulnerability. This significant decline suggests that the Binary Scheme cannot effectively handle paraphrase attacks due to its reliance on a score function as its statistical power. Distorted text can cause the score function to produce incorrect results when recalculating the secret key.

In contrast, our method incorporates fine-grained evaluation mechanisms for the secret key and modified text, which not only ensures strong performance under normal conditions but also maintains reliability against various user attacks. These results underscore the superiority of our scheme in practical, adversarial scenarios.

\begin{table}[h]
    \centering
    \small
    \begin{tabular}{l|ccc}
    \toprule
        & Basic & Translation & Paraphrase\\
        \midrule
        $e$ & 3.5 & 3.2 & 3.4 \\
        $e_\text{attack}$ & 4.3 & 7.3 & 13.4 \\
    \bottomrule
    \end{tabular}
    \caption{\textbf{Error escalation under different attack methods.} The table presents the total error ($e$) and the error after attacks ($e_{\text{attack}}$) for three scenarios: Basic, Translation, and Paraphrase. The results demonstrate a significant increase in errors caused by attack methods, highlighting their impact on the system's robustness.}
    \label{attack comparison}
\end{table}

Beyond the Llama model, our scheme demonstrated consistent performance across other models, including OPT, Gemma, and phi, before and after attacks. This consistency underscores its ability to generalize across different large language models, making it a robust and versatile solution for watermarking applications.

\subsection{Quality}
\begin{table*}
\centering
\small
\resizebox{1\linewidth}{!}{
\begin{tabular}{llccccccccccc}
\toprule

\multirow{2}{*}{Model}  & \multirow{2}{*}{Scheme} & \multicolumn{2}{c}{Open-Ended($\downarrow\%$)} & Semantic($\downarrow\%$) & \multicolumn{4}{c}{Long Answer QA($\downarrow\%$)}  & \multicolumn{4}{c}{Single Choice QA($\downarrow\%$)}              \\ \cmidrule(lr){3-4} \cmidrule(lr){5-5} \cmidrule(lr){6-9} \cmidrule(lr){10-13}  
       &                         & C4              & Story            & SimCSE                             & MATH              & GSM8K            & Hellaswag                & BFCL              & English      &Italian         & France             & German                              \\ \midrule
\multirow{2}{*}{Llama-3.2-1B} & ITS & 13.6 & 14.3 & 39.7 & 94.7 & 92.1 & 55.3 & 58.7 & 4.2 & 0.0 & 6.4 & 0.9  \\
& Ours & \textbf{12.9} & \textbf{14.0} & \textbf{32.7} & \textbf{8.7} & \textbf{4.3} & \textbf{2.3} & \textbf{0.0} & \textbf{0.0} & \textbf{0.0} & \textbf{0.0} & \textbf{0.0}    \\
\midrule
 
\multirow{2}{*}{OPT-1.3B} & ITS & 14.6 & 15.1 & 49.7 & 100.0 & 100.0 & 55.2 & 58.8 & 5.2 & 9.1 & 7.3 & 2.7  \\
& Ours & \textbf{12.4} & \textbf{13.1} & \textbf{43.6} & \textbf{3.7} & \textbf{8.3} & \textbf{2.9} & \textbf{0.0} & \textbf{0.0} & \textbf{0.0} & \textbf{0.0} & \textbf{0.0}    \\
\midrule

\multirow{2}{*}{Gemma-2B} & ITS & 17.3 & 18.1 & 51.3 & 94.6 & 83.8 & 56.1 & 55.6 & 4.5 & 8.6 & 3.3 & 12.2  \\
& Ours & \textbf{14.2} & \textbf{12.7} & \textbf{46.9} & \textbf{18.2} & \textbf{2.9} & \textbf{2.1} & \textbf{3.3} & \textbf{0.0} & \textbf{0.0} & \textbf{0.0} & \textbf{0.0}   \\
\midrule

\multirow{2}{*}{phi-2B} & ITS & 15.2 & 18.3 & 53.8  & 100.0 & 100.0 & 44.2 & 48.1 & 4.8 & 9.6 & 11.7 & 3.1  \\
& Ours & \textbf{13.2} & \textbf{14.2} & \textbf{49.1} & \textbf{16.7} & \textbf{10.0} & \textbf{3.7} & \textbf{3.2} & \textbf{0.0} & \textbf{0.0} & \textbf{0.0} & \textbf{0.0}   \\
\bottomrule
\end{tabular}
}
\caption{\textbf{Comparative evaluation of text quality across four different LLMs using various schemes.} The table presents performance metrics (\textdownarrow{} indicates lower is better) for open-ended tasks (C4, Story), semantic similarity (SimCSE), long-answer QA (MATH, GSM8K, Hellaswag, BFCL), and single-choice QA (English, Italian, French, German). Results compare the ITS scheme with the proposed approach (Ours), demonstrating improvements in key metrics across models Llama-3.2-1B, OPT-1.3B, Gemma-2B, and phi-2B.}
\label{text quality}
\end{table*}

We evaluate the quality of watermarked output to measure how much the mark degrades the utility of the output while maintaining the detectability. We build a suite of tasks that language models might be used for and compare the quality of watermarked outputs on these tasks to the quality without watermarking. We compare our scheme with $g$ in~\ref{g_bs} and the watermark generated by the original ITS scheme. In the experiments, we keep the total error $e$ under 1\%. For the open-ended tasks, we instruct GPT-4~\cite{achiam2023gpt} to score the pair of the watermarked text and original text from 0 to 10. We still use the continuous generation of the C4 dataset~\cite{raffel2020exploring} and generate stories given a same prompt in WrtingPrompt dataset~\cite{fan2018hierarchical}. To check the range of GPT-4 score changing, we calculate $(\text{score}_{\text{original}}-\text{score}_{\text{watermark}})/\text{score}_{\text{original}}$ for output of same length of 100. To check the semantic score, we use SimCSE~\cite{gao2021simcse} as the metric. We only use 100 sets of watermarked texts with the original text from WritingPrompt. We set original text as the standard to measure the semantic score of watermarks. Meanwhile, to check the ability of the watermark when facing long answer QA. We adapt four popular datasets: MATH~\cite{hendrycksmath2021}, GSM8K~\cite{cobbe2021training}, Hellaswag~\cite{zellers2019hellaswag}, BFCL~\cite{berkeley-function-calling-leaderboard}. Eventually, we validate multiple languages in MMLU dataset~\cite{hendrycks2020measuring}. These datasets are measured by accuracy. We calculate the degradation of scores by $(\text{acc}_{\text{original}}-\text{acc}_{\text{watermark}})/\text{acc}_{\text{original}}$

The experiment results are shown in table~\ref{text quality}. The GPT-4 scores reveal a consistent trend across all models and datasets: Our scheme consistently results in less degradation compared to ITS. The average degradation for our scheme across all models and datasets in the Open-Ended section is 13.34\%, while the average degradation for ITS is 15.81\%. 
Compared to the model evaluation, semantic information is more sensitive to the perturbations. Both watermarking schemes lead to a more significant degradation in semantic similarity across all models. Ours scheme performs better on SimCSE (43.08\% vs 48.63\%) since the leading tokens block are the same as the original output.

The results of long answer QA datasets also show a consistent trend across all models: ITS scheme degrades the text quality, particularly on tasks requiring deterministic answers. This is likely because ITS generates watermarked text based on a secret key, introducing randomness that disrupts task performance. In contrast, our scheme achieves higher accuracy than ITS and remains closer to the original model's performance. This improvement stems from the entropy threshold used in our watermarking scheme, which helps preserve output consistency while embedding the watermark.

MMLU datasets over different languages (English, Italian, French, German) notably show that both ITS and Our scheme perform relatively well compared to their accuracy on other benchmarks. This can be attributed to the nature of MMLU tasks, where outputs are restricted to fixed choices like 'A', 'B', or 'C'. In such cases, even though ITS relies on a secret key, the uniform distribution of the random variable $u$ from 0 to 1 gives a high probability of selecting the correct answer. Similarly, our scheme benefits from this structure, with its entropy-based threshold ensuring consistency while embedding the watermark. 

\subsection{Analysis of Samplings}
\label{analysis of samplings}
\begin{figure}[h]
    \centering
    \includegraphics[width=0.86\columnwidth]{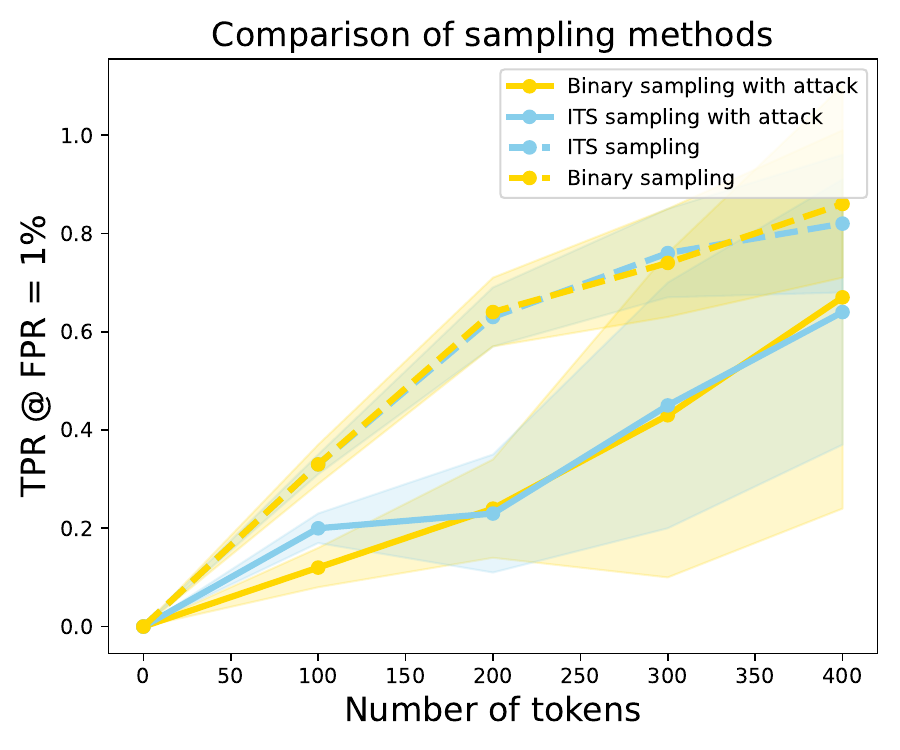}
    \caption{\textbf{Detectability of different sampling methods as text length increases.} The plot compares the True Positive Rate (TPR) at a fixed False Positive Rate (FPR = 1\%) for ITS sampling and Binary sampling, both with and without adversarial attacks. Results show that ITS sampling achieves higher detectability with shorter text lengths and maintains robustness under attacks, while Binary sampling demonstrates slower detectability growth and greater vulnerability to attacks as text length increases.}
    \label{function choosing}
\end{figure}

In this section, we analyze the detectability of two sampling functions in our watermarking scheme. We compare different $g$ functions as defined in eq.\ref{g_its} and eq.\ref{g_bs}, while keeping other parameters, such as $\lambda$, $T$, and the entropy function $\alpha$, constant. Additionally, we adapt the cost $d$ in eq.\ref{d_bs} for $g$ in eq.\ref{g_bs} to evaluate their performance. The detectability of watermarked text is measured using the True Positive Rate (TPR) when the False Positive Rate (FPR) is fixed at 1\%. Figure~\ref{function choosing} illustrates the detectability trends across different continuation lengths on the C4 dataset.

We find that the curves of the two sampling functions are closely aligned. By maintaining the FPR at a very low level (1\%), both sampling functions demonstrate detectability, particularly when the sequence length exceeds 200 tokens. To further evaluate the impact on text quality, we focused on the MATH and GSM8K datasets using the Llama model, as these datasets showed the most significant quality degradation. Unlike the previous text quality experiments, we employed multinomial sampling instead of top-$p$ sampling. The results, summarized in Table~\ref{sampling_ablation}, indicate comparable text quality across ITS, Binary, and Multinomial sampling methods, with minor differences observed across specific datasets.

\begin{table}[h]
    \centering
    \small
    \begin{tabular}{ccccc}
    \toprule
        Sampling & MATH & GSM8K &BFCL & Hellaswag\\ \midrule
        ITS & 13.1 & 22.3 & 10.5 &  29.1 \\ \midrule
        Binary & 12.3 & 19.4 & 10.7 &  30.2\\ \midrule
        Multinomial & 13.5 & 22.5 & 11.0 & 31.2\\\bottomrule
    \end{tabular}
    \caption{\textbf{Text quality evaluation of different sampling methods.} Performance is measured on MATH, GSM8K, BFCL, and Hellaswag datasets using the Llama model. The results indicate comparable quality across ITS, Binary, and Multinomial sampling, with minor differences in specific datasets.}
    \label{sampling_ablation}
\end{table}

\subsubsection{Samplings are equivalent}
\begin{figure}[h]
    \centering
    \includegraphics[width=0.8\columnwidth]{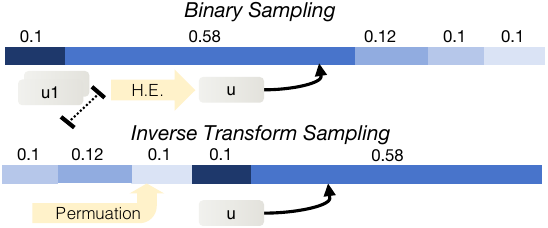}
    \caption{\textbf{Comparison of Binary Sampling and Inverse Transform Sampling.} The figure illustrates the mechanisms of the two sampling functions. Huffman Encoding (H.E.) is used in Binary Sampling to map a uniform random variable $u_1$ to a discrete binary outcome. In contrast, Inverse Transform Sampling applies a random permutation to introduce additional randomness while directly drawing $u$ from the uniform distribution $U[0,1]$.}
    \label{sampling analysis}
\end{figure}


From the experiments above, we observe that Binary Sampling (BS) and Inverse Transform Sampling (ITS) are equivalent to a certain extent. As shown in Figure~\ref{sampling analysis}, both methods depend on the secret key, which introduces controlled randomness into the sampling process.
In Binary Sampling, the secret key $u_1, \ldots, u_n$ is drawn from the uniform distribution $U[0,1]$, and each value is mapped to a discrete binary outcome ($0$ or $1$) using Huffman Encoding (H.E.). In contrast, Inverse Transform Sampling directly samples $u$ from the uniform distribution $U[0,1]$ and applies a random permutation to introduce additional randomness.

Compared to multinomial sampling, these two methods offer a more structured way to incorporate randomness, relying on the secret key to control the sampling process. This structured randomness ensures that the outputs from both BS and ITS exhibit similar behavior, thereby demonstrating their equivalence in practical applications.

\subsection{Generalization to Large Model}
To validate the effectiveness of our scheme on larger models, we adapt it to the Llama-3.1-8B model~\cite{dubey2024llama}. The evaluation is conducted on four challenging Long QA datasets: MATH, GSM8K, BFCL, and Hellaswag. The results are presented in Table~\ref{generalization_table}.

We observe that, as the model size increases, the degradation rate of the ITS scheme declines due to the enhanced capabilities of the larger LLM. However, our scheme consistently outperforms the ITS scheme, achieving substantially lower degradation rates across all datasets. For instance, on the GSM8K dataset, our scheme reduces the degradation rate from 63.9\% (ITS) to 4.3\%, and on Hellaswag, it eliminates degradation (0.0\%). These findings demonstrate the robustness and adaptability of our scheme when generalized to larger models, highlighting its effectiveness in maintaining high performance even on challenging datasets.
\begin{table}[h]
    \centering
    \small
    \begin{tabular}{ccccc}
        \toprule
        Scheme & MATH & GSM8K &BFCL & Hellaswag\\ \midrule
        ITS & 73.2 & 63.9 & 34.3 & 43.4 \\ \midrule
        Ours & \textbf{6.4} & \textbf{4.3} & \textbf{3.4} & \textbf{0.0}  \\ 
        \bottomrule
    \end{tabular}
    \caption{\textbf{Generalization results on the Llama-3.1-8B model.} Performance is evaluated on four challenging Long QA datasets (MATH, GSM8K, BFCL, Hellaswag). Our scheme significantly reduces degradation rates compared to the ITS scheme, demonstrating superior robustness on larger models.}
    \label{generalization_table}
\end{table}

\section{Related Work}
Current watermark algorithms without changing the structure of LLMs are worked on in the logit generation stage and token sampling. \cite{kirchenbauer2023watermark} introduced the first LLM watermarking technique based on logit modification. This method partitions the vocabulary into a red and green list at each token position, using a hash function that depends on the preceding token. A bias $\delta$ is applied to the logits of each token in the green list. \cite{christ2023undetectable} use Huffman encoding (The details are shown in appendix~\ref{huffman encoding}) to sample tokens from uniform distribution. At detection, they use a score function as the statistic to validate whether the text is watermarked. \cite{kuditipudi2023robust} proposed a watermarking method using a long pseudo-random number sequence, randomly selecting a starting position for each insertion to introduce randomness. During detection, they incorporate a soft edit distance (Levenshtein distance) to align text with the sequence, setting $k$ as the chunk length and selecting the chunk with the minimum cost as the final cost. This alignment-based strategy ensures robustness, as even if the text is cropped or altered, a single preserved watermarked block can trigger a low $p$-value. In this work, we utilize the previous two sampling functions in our scheme. Meanwhile, we adapt the covariance metric in~\cite{kuditipudi2023robust} for our detection.

\section{Conclusion}
Our work addresses achieving a robust and effective watermarking framework for LLMs during the sampling stage. Recognizing the need for an indistinguishable and reliably detectable watermark, we bridge the gap in existing research by proposing a novel approach grounded in mathematical consistency and validated through empirical performance. Our framework successfully capitalizes on the advantages of sampling-stage watermarking while mitigating its inherent trade-offs, ensuring high text quality and robust detection capabilities. This contribution not only advances the theoretical understanding of watermarking in generative models but also demonstrates practical viability, paving the way for more secure and reliable applications of LLMs.
\newpage

\bibliographystyle{named}
\bibliography{main}

\newpage
\appendix
\section{Collision by the analogy of birthday problem}
\label{birthday problem}
\textbf{The Problem Setup:}\\
Input: the secret key of length $n$, shift $\tau$\\
Output: the expectation of the number of tokens $m$\\
It's obvious to see there is a total of $k = \lfloor n/m\rfloor$ total independent secret key sequence.\\
To analogize the birthday problem, we have $k$ possible days and choose $l$ people. A collision occurs if two people share the same birthday.\\
$$
\P_{\text{no collision}}=\prod_{i=0}^{l-1}(1-\frac{i}{k})
$$
For large $k$ and $e^{-x}\approx1-x$, the equality approximate to:
$$
\P_{\text{no collision}}\approx \exp{(-\frac{l^2}{2k})}
$$
When $\P_{\text{collision}}$ is less than a probability $p$:
$$
\P_{\text{collision}}\approx 1-\exp{(-\frac{l^2}{2k})}\leq p
$$
Then we can have:
$$
l\leq \sqrt{k*(-2\ln(1-p))}
$$

\section{Indistinguishability of ITS}
\label{indis_its}
Suppose we have polynomial time queries $t=\text{poly}(\lambda)$. Let $r^{(1)}, r^{(2)}, \ldots r^{(t-1)}$ be the seed of responses $Y^{(1)}, Y^{(2)}, \ldots Y^{(t-1)}$. Considering for some $k\in[t]$ the watermarking algorithm stops before collecting enough entropy, we let $r^{(k)}\coloneq \operatorname{None}$. Define set $B\coloneq\{r^{(1)}, r^{(2)}, \ldots r^{(t-1)}\}\setminus \{\operatorname{None}\}$.
For any $r^{(k)}=\operatorname{None}$, it's trivial to show the indistinguishability since the text is sampled from the normal distribution. Then we will show $\P[r^{(t)}\in B]\leq \text{negl}(\lambda)$. Let $l^{(k)}$ denote the length of tokens made for seed $r^{(k)}$ and $y_i^{(k)}$ denote the tokens.
\begin{align*}
&\P[r^{(t)}\in B]\\
&=\P\left[r^{(t)}\in\{r^{(1)}, \ldots,r^{(t-1)}\}\setminus\{\operatorname{None}\}\right]\\
&\leq \sum_{k=1}^{t-1} \P[r^{(t)}=r^{(k)} \text{ and } r^{(t)}\neq \operatorname{None}]\\
&=\sum_{k=1}^{t-1}\mathds{1}\left[\sum_{i=1}^{l^{(k)}-1}1-p(y_i^{(k)})< \lambda\leq \sum_{i=1}^{l^{(k)}} 1-p(y_i^{(k)})\right]\\
&\prod_{i=1}^{l^{(k)}}p(y_i^{(k)})\\
&\leq \sum_{k=1}^{t-1}\mathds{1}\left[\lambda\leq \sum_{i=1}^{l^{(k)}} 1-p(y_i^{(k)})\right]\prod_{i=1}^{l^{(k)}}p(y_i^{(k)})\\
&= \sum_{k=1}^{t-1}\mathds{1}\left[\lambda\leq -\log\prod_{i=1}^{l^{(k)}}p(y_i^{(k)})\right]\prod_{i=1}^{l^{(k)}}p(y_i^{(k)})\\
&\leq (t-1)2^{-\lambda}\\
\end{align*}

\section{Binary construction}
\label{binary construction}
Now define the interval:
$$
I(Y)=[p(\{y:y<Y\}),p(\{y:y\leq Y\})].
$$
It's obvious to see:
$$
\mathbb{E}(\eta(Y))=\mathbb{E}(h)=\frac{1}{2}.
$$
For any event $I \subset [0,1]$ we have
\begin{align*}
\mathbb{P}(h\in I|Y)&=\frac{\mathbb{P}(h\in I, Y)}{\mu(Y)}\\
&=\frac{\mathbb{P}(Y|h\in I)\mathbb{P}(h\in I)}{\mu(Y)}\\
&=\frac{|I(Y)\cap I|}{I(Y)}
\end{align*}
Then we have 
\begin{align*}
\mathbb{E}(h|Y)&=\mathbb{E}\left[\mu\{y:y<Y\}+\frac{I(Y)}{2}\mid Y\right]\\
&=\frac{(Y-1)(1-p(Y))}{n-1}+\frac{p(Y)}{2}\\
&=\frac{1}{2}+(\eta(Y)-\frac{1}{2})(1-p(Y))
\end{align*}
For the covariance:
\begin{align*}
\operatorname{Cov}(h, \eta(Y))
&=\mathbb{E}\left[(h-\mathbb{E}(h))(\eta(Y)-\mathbb{E}(\eta(Y))\right]\\
&=(1-p(Y))\operatorname{Var}(\eta(Y))
\end{align*}
It's trivial to show $\mathbb{E}(d(Y, \xi^{(t)}))=0$ since $Y$ is independent to $\xi^{(t)}$. Thus we have,
\begin{align*}
\mathbb{E}\left[d(Y,\xi^{(t)})-d(Y, \xi)\right]
&=m\operatorname{Cov}(h, \eta(Y))\\
&=m\operatorname{Var}(\eta(Y))(1-p(Y))
\end{align*}

\section{Proof of p-value}
\label{proof_p}
By inserting the equation $$\mathbb{E}\left[d(Y, \xi^{(t)})-d(Y, \xi)\mid Y\right]=m\operatorname{Var}(\eta(Y))\alpha(Y)
$$ and Hoeffding's inequality, for $j \in [n]$ that 
\begin{align*}
&\P\left(d(Y_{i+1:i+k},\xi_{j+1:j+k}^{(t)}) \leq d(Y_{i:i+k-1},\xi_{i+1:i+k})\right)\\ 
& \leq \P\left(d(\wt{Y},\xi_{1:m}) - \Ep[d(\wt{Y},\xi_{1:m})] \geq k\operatorname{Var}(\eta(Y))\alpha/2\right) \\
& + \P\left(\Ep[d(\wt{Y},\xi_{j+1:j+m}')] - d(\wt{Y},\xi_{j+1:j+m}') \geq k\operatorname{Var}(\eta(Y)) \alpha/2\right) \\
& \leq 2\exp\left(-m \operatorname{Var}(\eta(Y))^2 \alpha^2/2\right).
\end{align*}

\section{Huffman encoding }
\label{huffman encoding}
The secret key $\xi$ shared by the watermarked model provider will be a sequence $\vec{u}=u_1, u_2, \ldots, u_m$, where each $u_i\sim U[0,1]$. To utilize this property, we follow the setting in \cite{christ2023undetectable}, they encode each token in $\mc{V}$ as a distinct string in $\{0, 1\}^{\log|\mc{V}|}$. Let $E$ denote the Huffman encoding function, and let $p_i$ be a distribution over $\mc{V}$ output by $M$. We convert $p_i$ into a series of distributions $p'_{i,j}$, where $j$ is the bit of $p_i$, and $p'_{i,j}$ is the binary distribution $\{0, 1\}$. 
\begin{algorithm}
\caption{Huffman encoding}
    \KwData{All token distributions $p_1, \ldots, p_{|\mc{V}|}$}
    \KwResult{Binary representations of all tokens $p_{i,1},\ldots,p_{i,\log|\mc{V}|}|_{i\in1,\ldots, |\mc{V}|}$}
    \For{$i\in \mc{V}$}{
        \For{$j\in\log|\mc{V}|$}{
            $p'_{i,j}(0) = \P[E(p_i)_j]$
            }
    }
\end{algorithm}

We encode each token $T$ in the vocabulary in the binary representation (bit tensor) before using.

\end{document}